\documentclass{article}

\usepackage{microtype}
\usepackage{times}
\usepackage{epsfig}
\usepackage{graphicx}
\usepackage{subfigure}
\usepackage{amsmath}
\usepackage{amssymb}
\usepackage{booktabs} 
\usepackage{array,multirow}
\usepackage{collcell}
\usepackage[dvipsnames,table]{xcolor}
\usepackage{fp}
\usepackage{xstring}
\usepackage{float}

\usepackage{xfrac}

\DeclareMathOperator*{\argmin}{argmin}
\DeclareMathOperator*{\cost}{cost}

\definecolor{LightRed}{rgb}{.99,.5,.5}
\definecolor{LightGreen}{rgb}{.5,.99,.5}
\newcommand{\checkvalue}[1]{
  \IfDecimal{#1}{
  \ifdim #1pt<0pt
    \FPset\inp{#1}
    \FPeval\oup{min(100,max(0,100*((10+\inp)/10)))}
    \xdef\oup{\oup}
    \cellcolor{white!\oup!LightRed}
    {{\hspace{-5pt}#1\%}}
  \else
    \FPset\inp{#1}
    \FPeval\oup{min(100,max(0,100*((\inp)/10)))}
    \xdef\oup{\oup}
    \cellcolor{LightGreen!\oup!white}
    {{\hspace{-5pt}#1\%}}
  \fi
  }
  {#1}
}

\newcommand{\checkfloat}[1]{
  \IfDecimal{#1}{
  \ifdim #1pt<0.65pt
    \FPset\inp{#1}
    \FPeval\oup{min(100,max(0,100*((\inp-.089)/.65)))}
    \xdef\oup{\oup}
    \cellcolor{white!\oup!LightRed}
    {{\hspace{-5pt}{#1}\%}}
  \else
    \FPset\inp{#1}
    \FPeval\oup{min(100,max(0,100*((\inp-0.65)/(1.915-0.65))))}
    \xdef\oup{\oup}
    \cellcolor{LightGreen!\oup!white} 
    {{\hspace{-5pt}{#1}\%}}
  \fi
  }
  {#1}
}

\newcolumntype{C}{>{\collectcell\checkvalue}c<{\endcollectcell}}
\newcolumntype{F}{>{\collectcell\checkfloat}c<{\endcollectcell}}

\usepackage{hyperref}

\usepackage[accepted]{icml2020}

\icmltitlerunning{Which Tasks Should Be Learned Together in Multi-task Learning?}

\begin{document}

\twocolumn[
\icmltitle{Which Tasks Should Be Learned Together in Multi-task Learning?}




\begin{icmlauthorlist}
\icmlauthor{Trevor Standley}{st}
\icmlauthor{Amir Zamir}{epfl}
\icmlauthor{Dawn Chen}{goo}
\icmlauthor{Leonidas Guibas}{st}
\icmlauthor{Jitendra Malik}{be}
\icmlauthor{Silvio Savarese}{st}
\url{http://taskgrouping.stanford.edu/}\vspace{-9pt}

\end{icmlauthorlist}

\icmlaffiliation{st}{Stanford University}
\icmlaffiliation{goo}{Google Inc.}
\icmlaffiliation{be}{The University of California, Berkeley}
\icmlaffiliation{epfl}{Swiss Federal Institute of Technology (EPFL)}

\icmlcorrespondingauthor{Trevor Standley}{tstand@cs.stanford.edu}

\icmlkeywords{Computer Vision, Multi-task Learning}

\vskip 0.3in
]


\begin{NoHyper}
\printAffiliationsAndNotice{}  
\end{NoHyper}

\begin{abstract}
Many computer vision applications require solving multiple tasks in real-time.
A neural network can be trained to solve multiple tasks simultaneously using \emph{multi-task learning}.
This can save computation at inference time as only a single network needs to be evaluated. Unfortunately, this often leads to inferior overall performance as task objectives can compete, which consequently poses the question: \textbf{which tasks should and should not be learned together in one network when employing multi-task learning?} 
We study task cooperation and competition in several different learning settings and propose a framework for assigning tasks to a few neural networks such that cooperating tasks are computed by the same neural network, while competing tasks are computed by different networks. Our framework offers a time-accuracy trade-off and can produce better accuracy using less inference time than not only a single large multi-task neural network but also many single-task networks.

\end{abstract}

\section{Introduction} \label{sec:intro}

Many applications, especially robotics and autonomous vehicles, are chiefly interested in using multi-task learning to reduce the inference time required to estimate many characteristics of visual input. 
For example, an autonomous vehicle may need to detect the location of pedestrians, determine a per-pixel depth, and predict objects' trajectories, all within tens of milliseconds.
In multi-task learning, multiple tasks are solved at the same time, typically with a single neural network. In addition to reduced inference time, solving a set of tasks jointly rather than independently can, in theory, have other benefits such as improved prediction accuracy, increased data efficiency, and reduced training time.

\begin{figure}
\begin{center}
\includegraphics[width=0.99\linewidth]{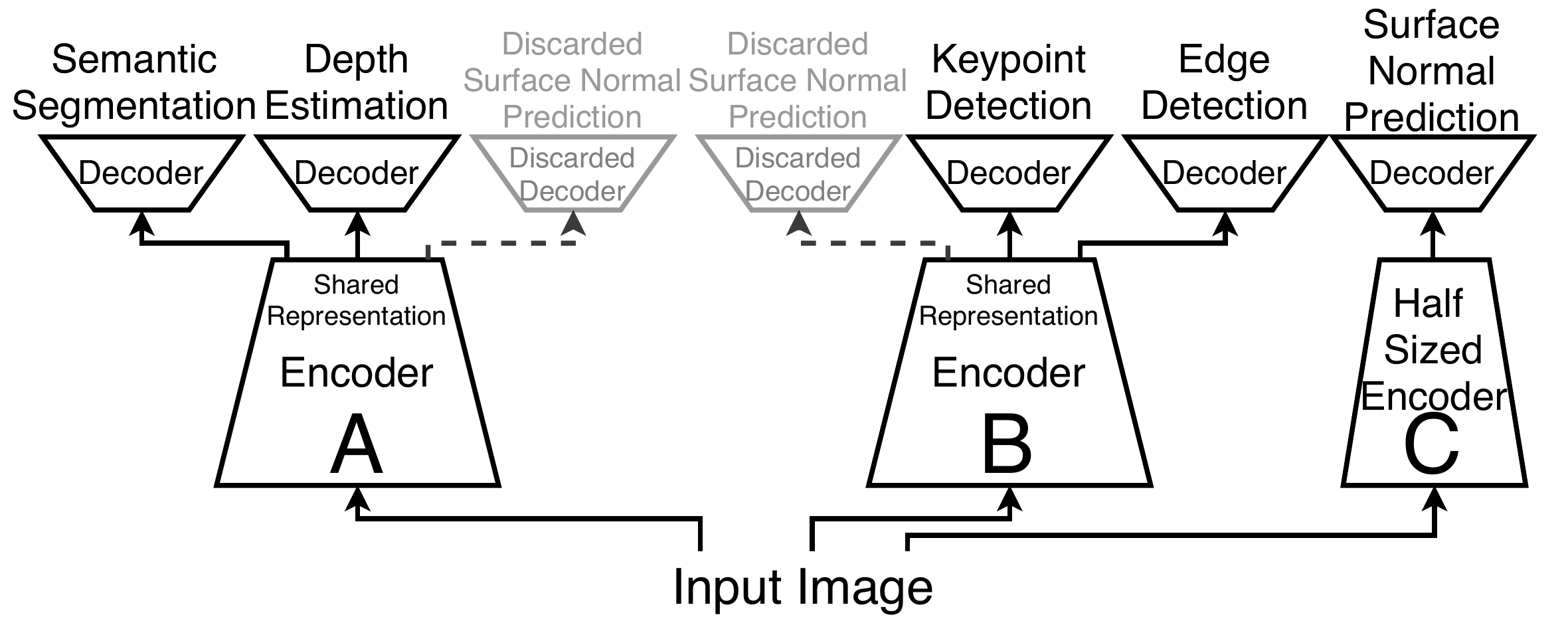}
\vspace{-10pt}
\caption{\footnotesize{\textbf{Given five example tasks to solve, there are many ways that they can be split into task groups for multi-task learning. How do we find the best one?} We propose a computational framework that, for instance, suggests the following grouping to achieve the lowest total loss, using a computational budget of 2.5 units: train network A to solve Semantic Segmentation, Depth Estimation, and Surface Normal Prediction; train network B to solve Keypoint Detection, Edge Detection, and Surface Normal Prediction; train network C with a less computationally expensive encoder to solve Surface Normal Prediction alone; including Surface Normals as an output in the first two networks were found advantageous for improving the other outputs, while the best Normals were predicted by the third network. This task grouping outperforms all other feasible ones, including learning \emph{all five tasks in one large network} or using \emph{five dedicated smaller networks}.}}
\label{fig:networks}
\vspace{-15pt}
\end{center}
\end{figure}

Unfortunately, the quality of predictions are often observed to suffer when a network is tasked with making multiple predictions due to a phenomenon called \textit{negative transfer}. In fact, multi-task performance can suffer so much that smaller independent networks are often superior. This may be because the tasks must be learned at different rates. Or because one task may dominate the learning leading to poor performance on other tasks. Furthermore, task gradients may interfere and multiple summed losses may make the optimization landscape more difficult.

On the other hand, when task objectives do not interfere much with each other, performance on both tasks can be maintained or even improved when jointly trained. Intuitively, this loss or gain of quality seems to depend on \textit{the relationship between the jointly trained tasks}. We empirically study these relationships in depth.

Prior work has studied the relationship between tasks for transfer learning \cite{taskonomy2018}. However, we find that transfer relationships are not highly predictive of multi-task relationships.
In addition to studying multi-task relationships, we attempt to determine how to produce good prediction accuracy under a limited inference time budget by assigning competing tasks to separate networks and cooperating tasks to the same network.

More concretely, this leads to the following problem: Given a set of tasks, $\mathcal{T}$, and a computational budget $b$ (e.g., maximum allowable inference time), what is the optimal way to assign tasks to networks with combined cost $\leq b$ such that a combined measure of task performances is maximized?

To this end, we develop a computational framework for choosing the best tasks to group together in order to have a small number of separate neural networks that completely cover the task set and that maximize task performance under a given computational budget. We make the intriguing observation that the inclusion of an additional task in a network can potentially improve the accuracy of the other tasks, even though the performance of the added task might be poor. This can be viewed as \emph{guiding} the loss of one task by adding an additional loss, for instance similar to curriculum learning~\cite{bengio2009curriculum}. 
Achieving this, of course, depends on picking the proper auxiliary task to be added -- our system can take advantage of this phenomenon, as schematically shown in Figure~\ref{fig:networks}. 



This paper has two main contributions. First, in Section~\ref{sec:relationships}, we provide an empirical study of a number of factors that influence multi-task learning including network size, dataset size, and how tasks influence one another when learned together. Then, in Section~\ref{sec:framework}, we outline a framework for assigning tasks to networks in order to achieve the best total prediction accuracy with a limited inference-time budget. We analyze the results and show that selecting the best assignment of tasks to groups is critical for good performance.

\section{Prior Work} \label{sec:prior_work}

\textbf{Multi-Task Learning:}
See \cite{DBLP:journals/corr/ZhangY17aa} for a good overview of multi-task learning. The authors of \cite{DBLP:journals/corr/Ruder17a} identify two clusters of contemporary techniques hard parameter sharing and soft parameter sharing. 


In hard parameter sharing, most or all of the parameters in the network are shared among all tasks. A known contemporary example of hard parameter sharing in computer vision is UberNet \cite{DBLP:conf/cvpr/Kokkinos17} which tackles 7 computer vision problems using hard parameter sharing. The author focuses on reducing the computational cost of training for hard parameter sharing, but experiences a rapid degradation in performance as more tasks are added to the network. Hard parameter sharing is also used in many other works such as  \cite{Thrun96islearning,Caruana1997,DBLP:journals/corr/abs-1809-04766,dvornik2017blitznet,NIPS2016_6393,pmlr-v70-pentina17a,doersch2017multitask,10.1007/978-3-319-46487-9_33,NIPS2017_6757,6378235,8123910,DBLP:conf/eccv/RuddGB16,leang2020dynamic,Chennupati_2019_CVPR_Workshops,suteu2019regularizing}.

Other works, such as \cite{NeurIPS2018_Sener_Koltun} and \cite{DBLP:conf/icml/ChenBLR18}, aim to dynamically re-weight each task's loss during training. The former work finds weights that provably lead to a Pareto-optimal solution, while the latter attempts to find weights that balance the influence of each task on network weights. \cite{kendall2017multi} uses uncertainty to weight tasks. \cite{8848395} and \cite{leang2020dynamic} compare loss weighting strategies. They both find no clear winner and similar performance among strategies, including a uniform weighting strategy. Similarly to us, they find that multi-task learning is often inferior to single task learning with multiple networks.

In soft or partial parameter sharing, either there is a separate set of parameters per task, or a significant fraction of the parameters are unshared. The models are tied together either by information sharing or by requiring parameters to be similar. Examples include  \cite{Dai2016InstanceAwareSS, duong-EtAl:2015:ACL-IJCNLP, Misra2016CrossStitchNF,Tessler:2017:DHA:3298239.3298465,36ebd0a9d3dd43f8bb0dbce896aa3c46,DBLP:conf/cvpr/LuKZCJF17}.

A canonical example of soft parameter sharing can be seen in  \cite{duong-EtAl:2015:ACL-IJCNLP}. The authors are interested in designing a deep dependency parser for languages such as Irish that do not have much treebank data available.
They leverage the abundant data from other languages and tie the weights of two networks together by adding an L2 distance penalty between corresponding weights and show substantial improvement for the target language. 

Another example of soft parameter sharing is Cross-stitch Networks  \cite{Misra2016CrossStitchNF}. Starting with separate networks for two tasks, the authors add `cross-stitch units' between them, which allow each network to peek at the other network's hidden layers.
This approach reduces but does not eliminate task interference, and the overall performance is less sensitive to the relative loss weights.

\cite{liu2019end} adds a per-task attention layer after each layer, and \cite{DBLP:conf/cvpr/ManinisRK19} adds task specific residual paths and adversarial training to match gradients. Both run the network separately for each task.

Finally, \cite{bingel-sogaard-2017-identifying} studies task interaction for NLP, and \cite{zamir2020consistency} makes use of task relationships to improve the predictions for related tasks.

Unlike our method, none of the aforementioned works attempt to discover good groups of tasks to train together. Also, soft parameter sharing does not reduce inference time, a major goal of ours.

Hybrid approaches such as PathNet \cite{DBLP:journals/corr/FernandoBBZHRPW17}, and AdaShare \cite{sun2019adashare} attempt to learn what to share and what not to. These works do not attempt to discover good pairs of source and target tasks nor to meet an inference time budget. Most importantly, they cannot achieve much inference-time speedup from module reuse.

\textbf{Transfer Learning:}
Transfer learning \cite{NIPS1992_641,10.1007/978-3-642-04180-8_52,silver2008guest, finn2016deep,mihalkova2007mapping,niculescu2007inductive,luolabel,Razavian:2014:CFO:2679599.2679731,5288526,DBLP:journals/corr/abs-1801-06519,DBLP:journals/corr/FernandoBBZHRPW17,DBLP:journals/corr/RusuRDSKKPH16} is similar to multi-task learning in that solutions are learned for multiple tasks. Unlike multi-task learning, however, transfer learning methods often assume that a model for a source task is given and then adapt that model to a target task. Transfer learning methods generally neither seek any benefit for source tasks nor a reduction in inference time as their main objective.




\textbf{Neural Architecture Search (NAS):}
Many recent works search the space of network architectures to find ones that perform well \citep{DBLP:conf/cvpr/ZophVSL18,45826,liu2018progressive,46637,xie2018snas,elsken2018efficient,pmlr-v97-zhou19e,baker2017designing,47354}. This is related to our work as we search the space of task groupings. Just as with NAS, the computationally found groupings often perform better than human-engineered ones. 

\textbf{Task Relationships:}
Our work is related to \textit{Taskonomy} \cite{taskonomy2018} which studied the relationships between visual tasks for \textit{transfer learning} and introduced a dataset with over 4 million images and corresponding labels for 26 tasks. A number of recent works further analyzed task relationships \cite{pal2019zeroshot,dwivedi2019,achille2019task2vec,wang2019neural} for transfer learning. 
While they extract relationships between these tasks for \emph{transfer learning}, we are interested in the \emph{multi-task learning} setting. Interestingly, we find notable differences between \emph{transfer task affinity} and \emph{multi-task affinity}. Their method also differs in that they are interested in labeled-data efficiency and not inference-time efficiency. Finally, the transfer quantification approach taken by Taskonomy (readout functions) is only capable of finding relationships between the high-level bottleneck representations developed for each task, whereas structural similarities between tasks at all levels are potentially relevant for multi-task learning.

\section{Experimental Setup}
\textbf{Dataset:} We perform our study using the Taskonomy dataset \cite{taskonomy2018}, which is currently the largest multi-task dataset for computer vision with diverse tasks. The data was obtained from 3D scans of about 600 buildings. The dataset has about 4 million examples, which we divided into about 3.9 million training instances (200k for Setting 3), about 50k validation instances, and about 50k test instances. There was no overlap in the buildings that appeared in the training and test sets.  

\textbf{Task Sets:} We have selected two sets of five tasks each from this dataset. \textbf{Task Set 1} includes \textit{Semantic Segmentation, Depth Estimation, Surface Normal Prediction, SURF Keypoint Detection}, and \textit{Canny Edge Detection}. One semantic task, two 3D tasks, and two 2D tasks are included. These tasks were chosen to be representative of major task categories, but also to have enough overlap in order to test the hypothesis that similar tasks will train well together. \textbf{Task Set 2} includes \textit{Auto Encoder, Surface Normal Prediction} again, \textit{Occlusion Edges, Reshading}, and \textit{Principal Curvature}. For a detailed definition of each of these tasks, see the Supplemental Materials for the Taskonomy paper. Cross-entropy loss was used for Semantic Segmentation, while an $L1$ loss was used for all other tasks. 

\textbf{Architectures:} In all experiments, we used a standard encoder-decoder architecture with a modified Xception \cite{chollet2017xception}) encoder. Our choice of architecture is not critical and was chosen for its reasonably low inference time. All max-pooling layers were replaced by $2\times2$ convolution layers with a stride of 2, similar to \cite{DBLP:conf/eccv/ChenZPSA18}. This network has 16.5 million parameters and requires 6.4 billion multiply-adds per image. The input image size for all networks is $256\times256$.

In order to study the effect of network size on task relationships, we also define a smaller \textbf{Xception17} network. For Xception17, the Xception network encoder was simplified to have 17 layers and the middle flow layers were reduced to having 512 rather than 728 channels, resulting in about 4 million parameters. This corresponds to 2.28 billion multiply-adds. 

Our decoders were designed to be lightweight, with four transposed convolutional layers \cite{7410535} and four convolutional layers.

\textbf{Settings:} We run our experiments under four settings.

\begin{table}[h]
\centering
 \scriptsize
\begin{tabular}{c|c|c|c|c}
& Network & Training Set Size & Task Set & Purpose\\
\hline
Setting 1 & \textbf{Xception17} & 3.9 million & Task Set 1 & Network Size\\
Setting 2 & Xception & 3.9 million & Task Set 1 & Control\\
Setting 3 & Xception & \textbf{200 thousand} & Task Set 1 & Dataset Size\\
Setting 4 & Xception & 3.9 million & \textbf{Task Set 2} & Task Set\\
\end{tabular}

\label{table:settings}
\end{table}

We test the effect of network size on multi-task learning with \textbf{Setting 1}. It uses a smaller and less deep network than the other settings.
\textbf{Setting 2} is the control. It uses a large network, the full dataset, and Task Set 1. We test the effect of dataset size on multi-task learning with \textbf{Setting 3}. Here we limit ourselves to only 200 thousand training instances.
Finally, we test other task relationships with \textbf{Setting 4}, which uses Task Set 2.

\textbf{Training Details:} All training was done using PyTorch \cite{paszke2017automatic} with Apex for fp16 acceleration \cite{micikevicius2017mixed}. 

The training loss we used was the unweighted mean of the losses for the included tasks. Networks were trained with an initial learning rate of 0.1, which was reduced by half every time the training loss stopped decreasing. Networks were trained until their validation loss stopped improving. The network with the highest validation loss (checked after each epoch of 20\% of our data) was saved. No hyper-parameter search was conducted.

\textbf{Comparison:} We define a computational network cost unit, the Standard Network Time (SNT). In Setting 1, an SNT is the number of multiply-adds in Xception17. In the other settings, an SNT is the number of multiply-adds in the normal Xception network. In order to produce our smaller/larger models to compare within each setting, we shrunk/grew the number of channels in every layer of the encoder such that it had the appropriate number of multiply-adds.

\textbf{Trained Networks:} In each setting, we train a 1-SNT network for each of the $2^n-1$ feasible subsets of our setting's 5 tasks. Thus, we train 5 single-task networks, 10 two-task networks, 10 three-task networks, 5 four-task networks, and a single five-task network. Another five single-task networks were trained, each having a half-size ($\sfrac{1}{2}$-SNT) encoder and a standard decoder. Finally, we trained a number of fractional-SNT single-task networks as comparisons and baselines.


\section{Study of Task Relationships} \label{sec:relationships}
We study a number of factors that influence a multi-task network's performance. We look at the relationships between tasks in each setting, and compare them to the relationships from our other settings. 

\textbf{Setting 1:} The smaller network, Xception17.

First, we analyze how the number of tasks included in a multi-task network affects performance. Table~\ref{table:perf_vs_number} shows the performance of multi-task learning relative to independent training. 

\setlength{\tabcolsep}{3pt}
\begin{table}[h!]
\centering
\scriptsize
\begin{tabular}{lCCCCC}
\toprule
& 1 task & 2 tasks & 3 tasks & 4 tasks & 5 tasks \\
\hline
\hline
1-SNT each & 0.00 & -7.56 & -7.23 & -10.69 & -19.00 \\
1-SNT total &0.00 & -1.15 & 4.23 & 4.86 & 0.34 \\

\bottomrule
\end{tabular}
\caption{\footnotesize{\textbf{Performance of multi-task learning \emph{relative} to independent (i.e. single-task) training.} For example, when a 4-task network is compared with four 1-SNT single-task networks, the 4-task network sees a 10.69\% worse total loss on average. When the same network is compared to four $\sfrac{1}{4}$-SNT networks, the total loss is 4.86\% better on average.}}
\vspace{-5pt}
\label{table:perf_vs_number}
\end{table}


\setlength{\tabcolsep}{2pt}
\begin{table}[h]
\centering
\scriptsize
\begin{tabular}{clCCCCC|C}
\toprule
 & & \multicolumn{6}{c}{Relative Performance On}  \\
& & SemSeg & Depth & Normals & Keypoints & Edges & Average \\
\hline
\hline
\parbox[t]{2mm}{\multirow{5}{*}{\rotatebox[origin=c]{90}{Trained With}}}& SemSeg & -- & -5.41 & -11.29 & -4.32 & -34.64 &  -13.92 \\
&Depth & 4.17 & -- & -3.55 & 3.49 & 3.76 &  1.97 \\
&Normals & 8.50 & 2.48 & -- & 1.37 & 12.33  & 6.17 \\
&Keypoints & 4.82 & 1.38 & -0.02 & -- & -5.26  & 0.23 \\
&Edges & 3.07 & -0.92 & -4.42 & 1.37 & -- &  -0.23 \\

\hline
& Average & 5.14 & -0.62 & -4.82 & 0.48 & -5.95 &  -1.15 \\
\bottomrule
\end{tabular}

\vspace{-5pt}
\caption{\footnotesize{\textbf{Pairwise multi-task relationships in Setting 1.} The table lists the performance of every task when trained as a pair with every other task. For instance, when SemSeg (Semantic Segmentation) is trained with Depth, SemSeg performs 4.17\% better than when SemSeg is trained alone on a $\sfrac{1}{2}$-SNT network.}}
\label{table:first_order}
\vspace{-5pt}
\end{table}

In Setting 1, multi-task networks do not compare favorably to multiple single-task networks that are each allowed the same computational budget as the single multi-task network. However, when the single-task networks are shrunk so that they have the same total budget as the multi-task network, multi-task networks with 3, 4, or 5 tasks outperform the single-task networks on average. Nevertheless, two-task networks still do not compare favorably. Table~\ref{table:first_order} gives a more detailed view of these ten two-task networks, showing the performance on each task when it is trained with each other task relative to when it is trained alone using a $\sfrac{1}{2}$-SNT network. We see that the Normals task helps the performance of every other task with which it is trained.

\setlength{\tabcolsep}{3pt}
\begin{table}[h!]
\centering
\scriptsize
\begin{tabular}{lCCCC}
\toprule
& Depth & Normals & Keypoints & Edges \\
\hline
\hline
SemSeg & -0.62 & -1.39 & 0.25 & -15.78 \\
Depth & & -0.54 & 2.43 & 1.42 \\
Normals & & & 0.67 & 3.95 \\
Keypoints & & & & -1.95 \\

\bottomrule
\end{tabular}

\caption{\footnotesize{\textbf{The \textit{multi-task} learning affinity between pairs of tasks for Setting 1.}  These values show the average change in the performance of two tasks when trained as a pair, relative to when they are trained separately.}}
\label{table:affinity}
\end{table}

\setlength{\tabcolsep}{3pt}
\begin{table}[h!]
\centering
\scriptsize
\begin{tabular}{lFFFF}
\toprule
& Depth & Normals & Keypoints & Edges \\
\hline
\hline
SemSeg & 1.740 & 1.828 & 0.723 & 0.700 \\
Depth & & 1.915 & 0.406 & 0.468 \\
Normals & & & 0.089 & 0.118 \\
Keypoints & & & & 0.232 \\
\bottomrule
\end{tabular}
\caption{\footnotesize{\textbf{The \textit{transfer} learning affinities} between pairs of tasks according to the authors of Taskonomy \cite{taskonomy2018}. Forward and backward transfer affinities are averaged.}}
\label{table:trans_affinity}
\end{table}

In order to determine the between-task affinity for multi-task learning, we took the average of our first-order relationships matrix (Table~\ref{table:first_order}) and its transpose. The result is shown in Table~\ref{table:affinity}.  The pair with the highest affinity by this metric are Surface Normal Prediction and 2D Edge Detection. Quite surprisingly, our two 3D tasks, Depth Estimation and Surface Normal Prediction, do not score highly on this similarity metric. This contrasts with the findings for transfer learning in Taskonomy (Table~\ref{table:trans_affinity}), for which they have the highest affinity. In fact, there seems to be no correlation between multi-task and transfer learning affinities (tables ~\ref{table:affinity} and ~\ref{table:trans_affinity}) in this setting (Pearson's $r$ is $-0.12$, $p = 0.74$) .




\textbf{Setting 2:} The control setting.


We test the effect of network capacity on multi-task affinity, by retraining all of our networks using a higher-capacity encoder. We used the full Xception network \cite{chollet2017xception} this time, which uses 6.4 billion multiply-adds. The resulting data allows us to generate a new version of Table~\ref{table:first_order} for Setting 2 as Table~\ref{table:first_order_bigger}.

\setlength{\tabcolsep}{2pt}
\begin{table}[h]

\centering
\scriptsize
\begin{tabular}{clCCCCC|C}
\toprule
 & & \multicolumn{6}{c}{Relative Performance On}  \\
& & SemSeg & Depth & Normals & Keypoints & Edges & Average \\
\hline
\hline
\parbox[t]{2mm}{\multirow{5}{*}{\rotatebox[origin=c]{90}{Trained With}}}& SemSeg & -- & 3.00 & -2.79 & -5.20 & 27.80 & 5.70 \\
& Depth & 1.72 & -- & 1.18 & -3.52 & 25.73 & 6.28 \\
& Normals & 10.81 & 7.12 & -- & 88.98 & 71.59 & 44.62 \\
& Keypoints & 3.12 & -0.41 & -10.12 & -- & 61.07 & 13.42 \\
& Edges & 0.03 & -1.40 & -4.78 & -3.05 & -- & -2.30 \\
\hline
& & 3.92 & 2.08 & -4.13 & 19.30 & 46.54 & 13.54 \\
\bottomrule
\end{tabular}

\vspace{-5pt}
\caption{\footnotesize{\textbf{Pairwise multi-task relationships in Setting 2.}}}
\label{table:first_order_bigger}
\vspace{-5pt}
\end{table}

We can see by comparing Table~\ref{table:first_order_bigger} with Table~\ref{table:first_order} that tasks are much more likely to benefit from being trained together when using the larger network. However, there are still tasks that both suffer when trained together. Furthermore, the values in Table~\ref{table:first_order_bigger} do not seem to correlate with the values in Table~\ref{table:first_order} (Pearson's $r=0.08$). The affinities also do not seem to correlate with Taskonomy's transfer affinities (Pearson's $r=-0.14$). These results stress the necessity of using a automatic framework each particular setup to determine which tasks to train together.

\textbf{Setting 3:} Using only 5\% of the available training data.

We study the effect of training set size by using only 199,498 training instances (420 from each training building). This amount of data is more similar to other multi-task datasets in the literature. One common assumption is that multi-task learning is likely to be better in low-data scenarios, as MTL effectively allows you to pool supervision. However, we see this assumption violated in Table~\ref{table:first_order_less_data}, as most tasks suffer when trained with another task in this setting, especially when compared to Setting 2. Furthermore, were it not for the huge gains in the \textit{Edges} task in a couple of cases, MTL would be deleterious on average.

\setlength{\tabcolsep}{2pt}
\begin{table}[h]
\centering
\scriptsize
\begin{tabular}{clCCCCC|C}
\toprule
 & & \multicolumn{6}{c}{Relative Performance On}  \\
& & SemSeg & Depth & Normals & Keypoints & Edges & Average \\
\hline
\hline
\parbox[t]{2mm}{\multirow{5}{*}{\rotatebox[origin=c]{90}{Trained With}}}
& SemSeg & -- & 1.91 & -6.00 & -9.91 & -21.93 & -8.98 \\
& Depth & -12.63 & -- & 2.95 & 1.44 & -9.70 & -4.48 \\
& Normals & 8.32 & 15.38 & -- & -1.35 & 52.08 & 18.61 \\
& Keypoints & -5.84 & -7.21 & -2.26 & -- & 55.63 & 10.08 \\
& Edges & -5.62 & 6.02 & -4.16 & -5.02 & -- & -2.20 \\
\hline
& & -3.95 & 4.03 & -2.37 & -3.71 & 19.02 & 2.6 \\
\bottomrule
\end{tabular}

\vspace{-5pt}
\caption{\footnotesize{\textbf{Pairwise multi-task relationships in Setting 3.}}}
\label{table:first_order_less_data}
\vspace{-5pt}
\end{table}

Although we did not find correlations between the task relationships for the low-capacity model and those for the high-capacity model, the low-data relationships show a positive correlation with both. The small-data relationships correlate with the low-capacity relationships (Pearson's $r=+0.375, p=0.10$) as well as the high-capacity relationships (Pearson's $r=+0.558, p=0.01$). However, these affinities still do not correlate highly with Taskonomy's transfer-learning affinities (Pearson's $r=-0.235, p=0.51$).

\textbf{Setting 4:} Using Task Set 2.

Aside from the Auto Encoder task, the other four tasks in Task Set 2 nearly all have a moderate positive influence on one another. These four tasks are all very similar 3D tasks, so it seems that similar tasks can work well with one another in some situations. Unfortunately, the Auto Encoder task hurts the performance of the other tasks on average. Once again, we find no correlation with Taskonomy's transfer affinity (Pearson's $r=-0.153, p=0.674$).

\setlength{\tabcolsep}{2pt}
\begin{table}[h]
\centering
\scriptsize
\begin{tabular}{clCCCCC|C}
\toprule
 & & \multicolumn{6}{c}{Relative Performance On}  \\
& & AutoEnc & Normals & Occ Edges & Reshading & Curvature & Average \\
\hline
\hline
\parbox[t]{2mm}{\multirow{5}{*}{\rotatebox[origin=c]{90}{Trained With}}}
& AutoEnc & -- & -3.23 & -2.66 & 0.10 & -1.39  & -1.79 \\
& Normals & 19.31 & -- & 3.16 & 4.60 & 1.95  & 7.25 \\
& Occ Edges & 35.83 & -0.25 & -- & 1.15 & 0.84  & 9.39 \\
& Reshading & -24.46 & 3.71 & 3.16 & -- & 1.88  & -3.93 \\
& Princ Curv & 10.69 & 2.61 & 2.46 & 3.15 & --  & 4.73 \\
\hline
& & 10.34 & 0.71 & 1.53 & 2.25 & 0.82 & 3.13 \\
\bottomrule
\end{tabular}

\vspace{-5pt}
\caption{\footnotesize{\textbf{Pairwise multi-task relationships in Setting 4.}}}
\label{table:first_order_different_tasks}
\vspace{-5pt}
\end{table}

\textbf{Key Takeaways:} Ideally, the relationships between tasks would be independent of the learning setup. We find that this is not the case, therefore using a computational approach like ours for finding task affinities and groupings seems necessary. We saw that both network capacity and the amount of training data influence multi-task task affinity. We also found that more similar tasks don't necessarily train better together. We also find no correlation between multi-task task affinity and transfer learning task affinity in any setting.

Finally, the Normals task seems to improve the performance of the tasks it is trained with. In 15 out of 16 models that were trained with Normals, the other task improved. Furthermore, this improvement was better than the effect of co-training with any different task 13 out of 15 times. This may be because Normals have uniform values across surfaces and preserve 3D edges. However, the Normals task itself tends to perform worse when trained with another task.

We see that choosing which tasks to learn together is critical for achieving good performance. Now, we turn to the study of how to find the best tasks to train together.

\section{Task Grouping Framework} \label{sec:framework}


\textbf{Overview:} Our goal is to find a set of networks, each of which is trained on a subset of the tasks, that results in the best overall loss within a given computational budget. We do this by considering the space of all possible task subsets, training a network for each subset, and then using each network's performance to choose the best networks that fit within the budget. Because fully training a network for each subset may be prohibitively expensive, we outline two strategies for predicting training outcomes as well (Sec.~\ref{sec:approx}).



\textbf{Formal Problem Definition:} We want to minimize the overall loss on a set of tasks $\mathcal{T}=\{t_1,t_2,...,t_k\}$ given a limited inference time budget, $b$, which is the total amount of time we have to complete all tasks. Each neural network that solves some subset of $\mathcal{T}$ and that could potentially be a part of the final solution is denoted by $n$. It has an associated inference time cost, $c_n$, and a loss for each task, $\mathcal{L}(n,t_i)$ (which is $\infty$ for each task the network does not attempt to solve). A solution $\boldsymbol{S}$ is a set of networks that together solve all tasks. The computational cost of a solution is $\cost(\boldsymbol{S})=\sum_{n \in \boldsymbol{S}} c_n$. The loss of a solution on a task, $\mathcal{L}(\boldsymbol{S},t_i)$, is the lowest loss on that task among the solution's networks\footnote{In principle, it may be possible to create an even better-performing ensemble when multiple networks solve the same task, though we do not explore this.},  $\mathcal{L}(\boldsymbol{S},t_i)=\min_{n\in \boldsymbol{S}} \mathcal{L}(n,t_i)$. The overall performance for a solution is $\mathcal{L}(\boldsymbol{S})=\sum_{t_i\in\mathcal{T}}\mathcal{L}(\boldsymbol{S},t_i)$.

We want to find the solution with the lowest overall loss and a cost that is under our budget, $\boldsymbol{S}_b=\argmin_{\boldsymbol{S} : \cost(\boldsymbol{S})\leq b} \mathcal{L}(\boldsymbol{S})$. 

\subsection{Which Candidate Networks to Consider?} \label{ssec:network_training}

For a given task set $\mathcal{T}$, we wish to determine not just how well each \textit{pair} of tasks performs when trained together, but also how well each \textit{combination} of tasks performs together so that we can capture higher-order task relationships. To that end, our candidate set of networks contains all $2^{ \vert \mathcal{T}\vert}-1$ possible groupings: $\binom{\vert \mathcal{T}\vert}{1}$ networks with one task, $\binom{\vert \mathcal{T}\vert}{2}$ networks with two tasks,$\binom{\vert \mathcal{T}\vert}{3}$ networks with three tasks, etc. For the five tasks we use in our experiments, this is 31 networks, of which five are single-task networks.

The size of the networks is another design choice, and to somewhat explore its effects we also include 5 single task networks each with half of the computational cost of a standard network. This brings our total up to 36 networks.

In principle, one could include additional candidate networks of arbitrarily varying computational cost, and let the framework below decide which are worth using. It may even be advantageous to throw in networks trained with different architectures, task weights or training strategies. We do not explore this.



\subsection{Network Selection} \label{ssec:network selection}

Consider the situation in which we have an initial candidate set $\boldsymbol{C}_0=\{n_1,n_2,...,n_m\}$ of \textbf{fully-trained }networks that each solve some subset of our task set $\mathcal{T}$. Our goal is to choose a subset of $\boldsymbol{C}_0$ that solve all the tasks with total inference time under budget $b$ and the lowest overall loss. More formally, we want to find a solution $\boldsymbol{S}_b=\argmin_{\boldsymbol{S} \subseteq \boldsymbol{C}_0: \cost(\boldsymbol{S})\leq b} \mathcal{L}(\boldsymbol{S})$.

It can be shown that solving this problem is NP-hard in general (reduction from \textsc{Set-Cover}). However, many techniques exist that can optimally solve \textit{most} reasonably-sized instances of problems like these in acceptable amounts of time. All of these techniques produce the same solutions. We chose to use a branch-and-bound-like algorithm for finding our optimal solutions (pseudo code for the algorithm is in the supplemental material, and our implementation is on GitHub), but in principle the exact same solutions could be achieved by other optimization methods, such as encoding the problem as a binary integer program (BIP) and solving it in a way similar to Taskonomy \cite{taskonomy2018}.


Most contemporary MTL works use fewer than 4 unique task types, however, using synthetic inputs, we found that our branch-and-bound like approach requires less time than network training for all $2^{ \vert \mathcal{T}\vert}-1+\vert \mathcal{T}\vert$ candidates for fewer than ten tasks.

\subsection{Approximations for Reducing Training Time}\label{sec:approx}

This section describes two techniques for reducing the training time required to obtain a collection of networks as input to the network selection algorithm. Our goal is to produce task groupings with results similar to the ones produced by the complete search, but with less training time burden. Both techniques involve predicting the performance of a network without actually training it to convergence. The first technique involves training each of the networks for a short amount of time, and the second involves inferring how networks trained on more than two tasks will perform based on how networks trained on two tasks perform.

\subsubsection{Early Stopping Prior to Convergence} \label{sssec:dontwait}
We found a moderately high correlation (Pearson's $r = 0.49$) between the validation loss of our 36 candidate neural networks in Setting 1 after a pass through just 20\% of our data and the final validation loss of the fully trained networks. This implies that the task relationship trends stabilize early. We find that we can get decent results by running network selection on the lightly trained networks, and then simply training the chosen networks to convergence. This is a common technique in hyperparameter optimization, \cite{10.5555/2832581.2832731,10.5555/3122009.3242042}.

For our setup, this technique reduces the training time burden by about \textbf{20x} over fully training all candiate networks. Obviously, this technique does come with a prediction accuracy penalty. Because the correlation between early network performance and final network performance is not perfect, the decisions made by network selection are no longer guaranteed to be optimal once networks are trained to convergence. We call this approximation the Early Stopping Approximation (ESA) and present the results of using this technique in Section~\ref{sec:results1}.

\subsubsection{Predict Higher-Order From Lower-Order} \label{sssec:higher_order}
Do the performances of a network trained with tasks $A$ and $B$, another trained with tasks $A$ and $C$, and a third trained with tasks $B$ and $C$ tell us anything about the performance of a network trained on tasks $A$, $B$, and $C$?
As it turns out, the answer is yes. Although this ignores complex task interactions and nonlinearities, a simple average of the first-order networks' accuracies was a good indicator of the accuracy of a higher-order network. For example, if you have networks, a\&b with losses 0.1\&0.2, b\&c with 0.3\&0.4, and a\&c with 0.5\&0.6, the per-task loss estimate for a network with a\&b\&c would be $a=(0.1+0.6)/2=0.35$, $b=(0.2+0.3)/2=0.25$ and $c=(0.4+0.6)/2=0.5$.


Using this strategy, we can predict the performance of all networks with three or more tasks using the performance of all of the fully trained two task networks. First, simply train all networks with two or fewer tasks to convergence. Then predict the performance of higher-order networks, run network selection on both the trained and the predicted networks, then train the higher order networks from scratch.


With our setup, this strategy saves only about 45\% of the training time, compared with 95\% for the early stopping approximation, and it still comes with a prediction quality penalty. However, this technique requires only a quadratic number of networks to be trained rather than an exponential number, and would therefore theoretically win out when the number of tasks is large.

We call this strategy the Higher Order Approximation (HOA), and present its results in Section~\ref{sec:results1}.




\vspace{-2pt}
\section{Task Grouping Evaluation} \label{sec:results1}
\vspace{-1pt}
We applied our framework and approximations in all four settings. We computed solutions for inference time budgets from 1 SNT to 5 SNT at increments of $\sfrac{1}{2}$ SNT. The performance scores used for network selection were calculated on the validation set. Each solution chosen was evaluated on the test set. Note that the total loss numbers reported cannot be properly compared between settings because of small differences in label normalization and loss definitions.

\textbf{Baselines:} We compared our results with conventional methods, such as five single-task networks and a single network with all tasks trained jointly.

For Setting 1, we also compared with two multi-task methods in the literature. The first one is \cite{NeurIPS2018_Sener_Koltun}. We found that their algorithm under-weighted the Semantic Segmentation task too aggressively, leading to poor performance on the task and poor performance overall compared to a simple sum of task losses. We speculate that this is because the loss for Semantic Segmentation behaves differently from the other losses. Next we compared to GradNorm \cite{DBLP:conf/icml/ChenBLR18}. GradNorm's results were also slightly worse than classical MTL with uniform task weights, though that may be because we evaluated with uniform task weights, while GranNorm optimized for task weights that it computed. In any event, these techniques are orthogonal to ours and can be used in conjunction for situations in which they lead to better solutions than simply summing losses.

Each baseline was evaluated with multiple encoder sizes (more or fewer channels per layer) so that the results could be compared at many inference time budgets.


Finally, we compared our results to two control baselines illustrative of the importance of making good choices about which tasks to train together, `Random' and `Pessimal.' `Random' is a solution consisting of valid random task groupings that solve our five tasks. The reported values are the average of one million random trials. `Pessimal' is a solution in which we choose the networks that lead to the worst overall performance, though the solution's performance on each task is still the best among the networks that solve that task.

\begin{figure}[ht]

\begin{center}

\includegraphics[width=0.99\linewidth]{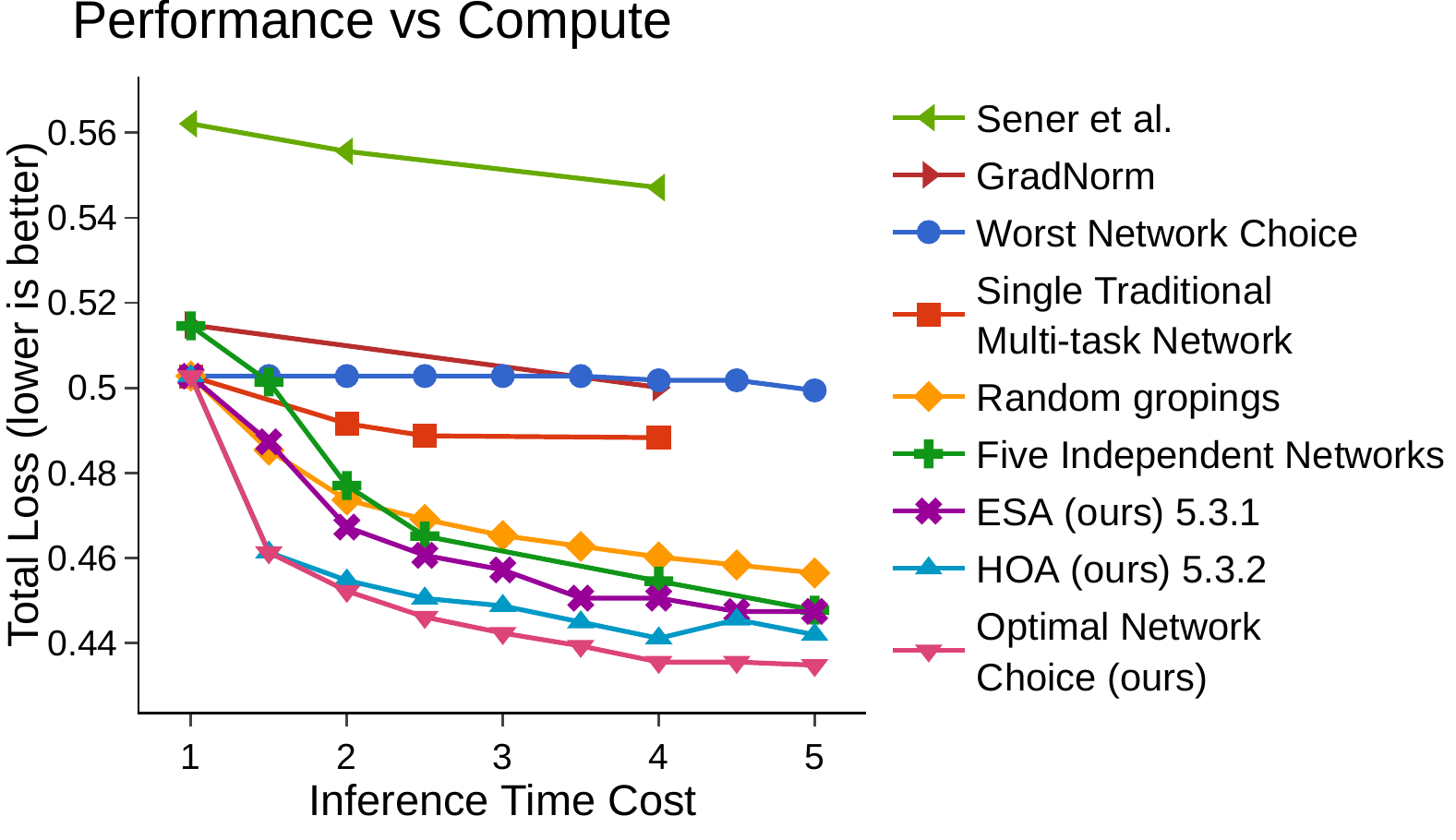}
\vspace{-10pt}
\caption{\footnotesize{\textbf{Performance/inference time trade-off for various methods in Setting 1.} We do not report error bars because the test set is large enough that standard errors are too small to be shown.}}
\label{fig:main_results}
\vspace{-10pt}
\end{center}
\end{figure}

\begin{figure}[ht]
\begin{center}
\vspace{-5pt}
\includegraphics[width=0.485\textwidth]{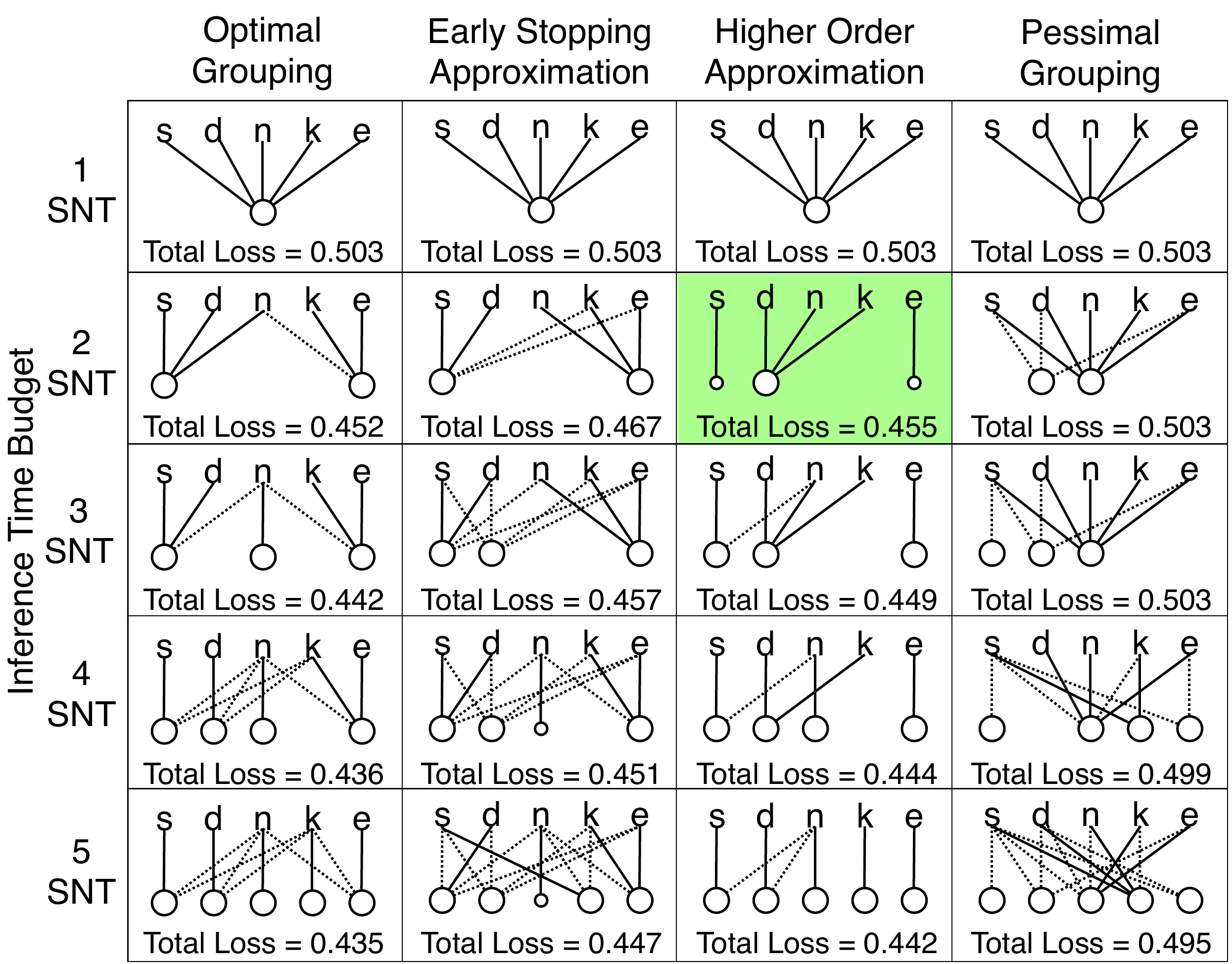}
\vspace{-10pt}
\caption{\footnotesize{\textbf{The task groups picked by each of our techniques for integer budgets between 1 and 5.} Networks are shown as {\Large $\circ$} (full-size) or $\circ$ (half-size). Networks are connected to the tasks for which they compute predictions. \textit{s: Semantic Segmentation, d: Depth Estimation, n: Surface Normal Prediction, k: Keypoint Detection, e: Edge Detection}. Dotted edges represent unused decoders. For example, the \textcolor{green}{green} highlighted solution consists of two half-size networks and a full-size network. The full-size network solves Depth Estimation, Surface Normal Prediction, and Keypoint Detection. One half-size network solves Semantic Segmentation and the other solves Edge Detection. The total loss for all five tasks is 0.455. The groupings for fractional budgets are shown in the supplemental material.}}
\label{fig:solutions}
\vspace{-10pt}
\end{center}
\end{figure}

\textbf{Setting 1:} Figure~\ref{fig:solutions} shows the task groups that were chosen for each technique, and Figure~\ref{fig:main_results} shows the performance of these groups along with those of our baselines. We can see that each of our methods outperforms the traditional baselines for every computational budget. 

When the computational budget is only 1 SNT, all of our methods must select the same model---a traditional multi-task network with a 1 SNT encoder and five decoders. This strategy outperforms GradNorm \cite{NeurIPS2018_Sener_Koltun}, as well as independent networks. However, solutions that utilize multiple networks outperform this traditional strategy for every budget $>$ 1.5---better performance can always be achieved by grouping tasks according to their compatibility.


When the computational budget is effectively unlimited (5 SNT), our optimal method picks five networks, each of which is used to make predictions for a separate task. However, three of the networks are trained with three tasks each, while only two are trained with one task each. This shows that the networks learned through multi-task learning were found to be best for three of our tasks (s, d, and e), whereas two of our tasks (n and k) are best solved individually. 

We also see that our optimal technique using 2.5 SNT and our Higher Order Approximation using 3.5 SNT can both outperform individual networks using 5 SNT total.



\begin{figure}[h]
\begin{center}
\includegraphics[width=0.99\linewidth]{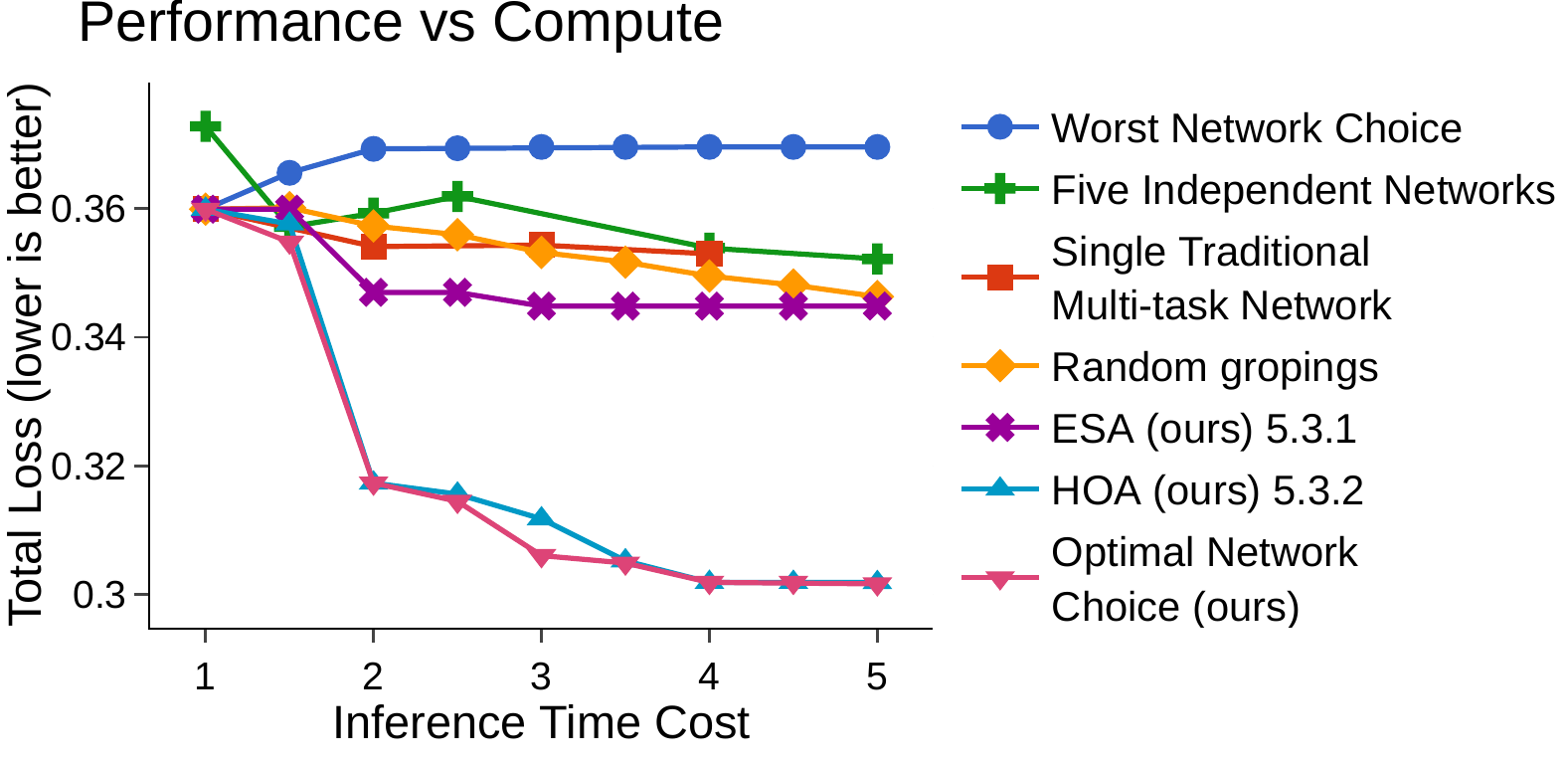}
\vspace{-10pt}
\caption{\footnotesize{\textbf{Performance/inference time trade-off in Setting 2.}}}
\vspace{-10pt}
\label{fig:big_results}
\end{center}
\end{figure}

\textbf{Setting 2:} When we apply our network selection framework on the performance data from this high-capacity network (Figure~\ref{fig:big_results}) we again see that our method outperforms both training an individual network for each task, as well as training all tasks together. This is true even though pairs of tasks tend to cooperate better in this setting. In fact, the performance of our groupings is superior by an even wider margin here. It should be noted that although ESA outperforms the baselines, there is a significant gap between ESA and the optimal solution. Perhaps stopping after training on more of the data would improve ESA's results.   

\begin{figure}[h]
\begin{center}
\includegraphics[width=0.99\linewidth]{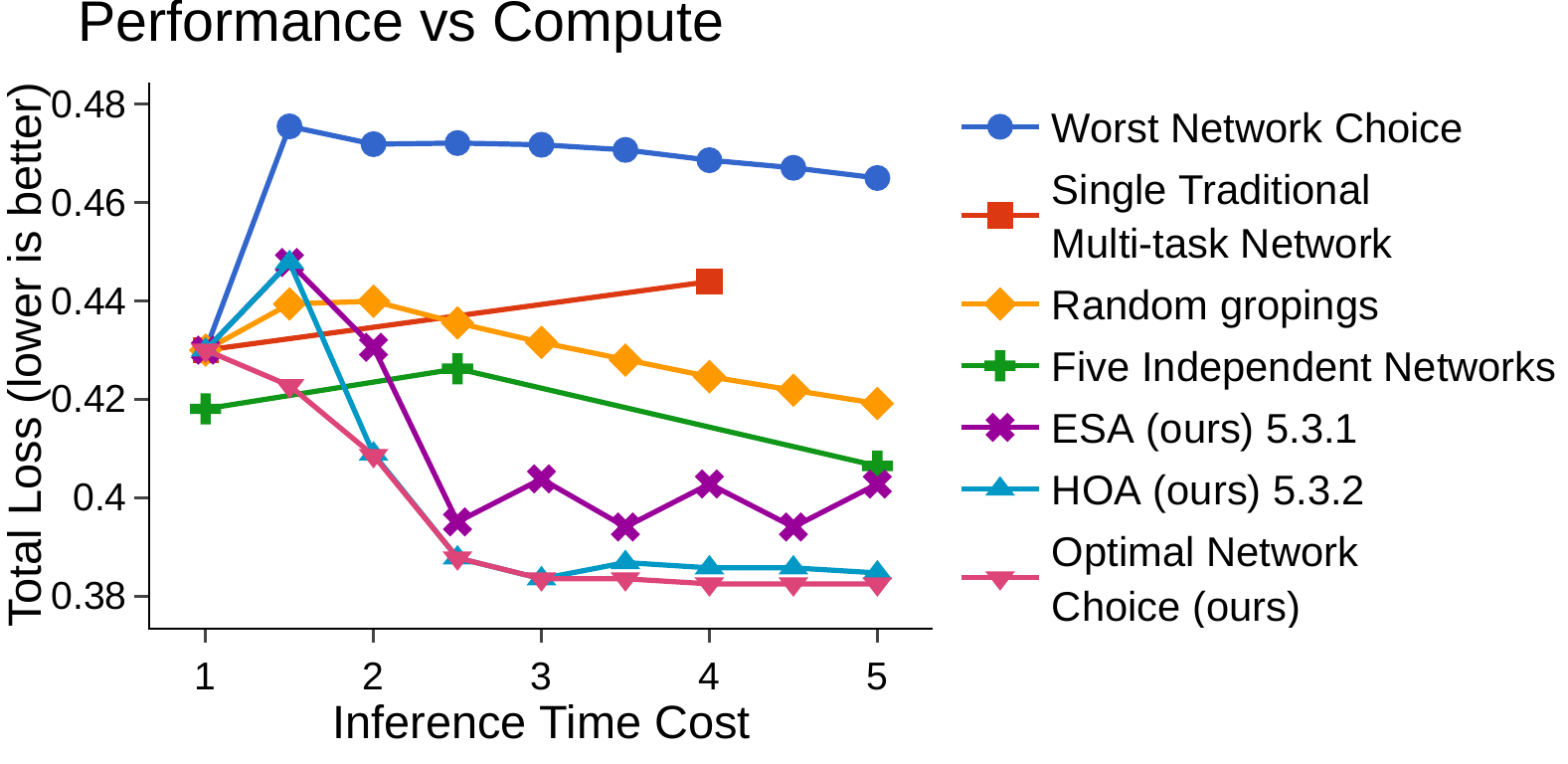}
\vspace{-10pt}
\caption{\footnotesize{\textbf{Performance/inference time trade-off in Setting 3.}}}
\vspace{-10pt}
\label{fig:small_data_results}
\end{center}
\end{figure}

\textbf{Setting 3:} We can see in Figure~\ref{fig:small_data_results} that our selection framework on the networks trained with only 200k examples is again superior to the baselines. For ESA in this setting, we ran through the entire 200k examples four times, rather than through only 20\% once. This represents the same amount of training as in the other settings.

\begin{figure}[h]
\begin{center}
\includegraphics[width=0.99\linewidth]{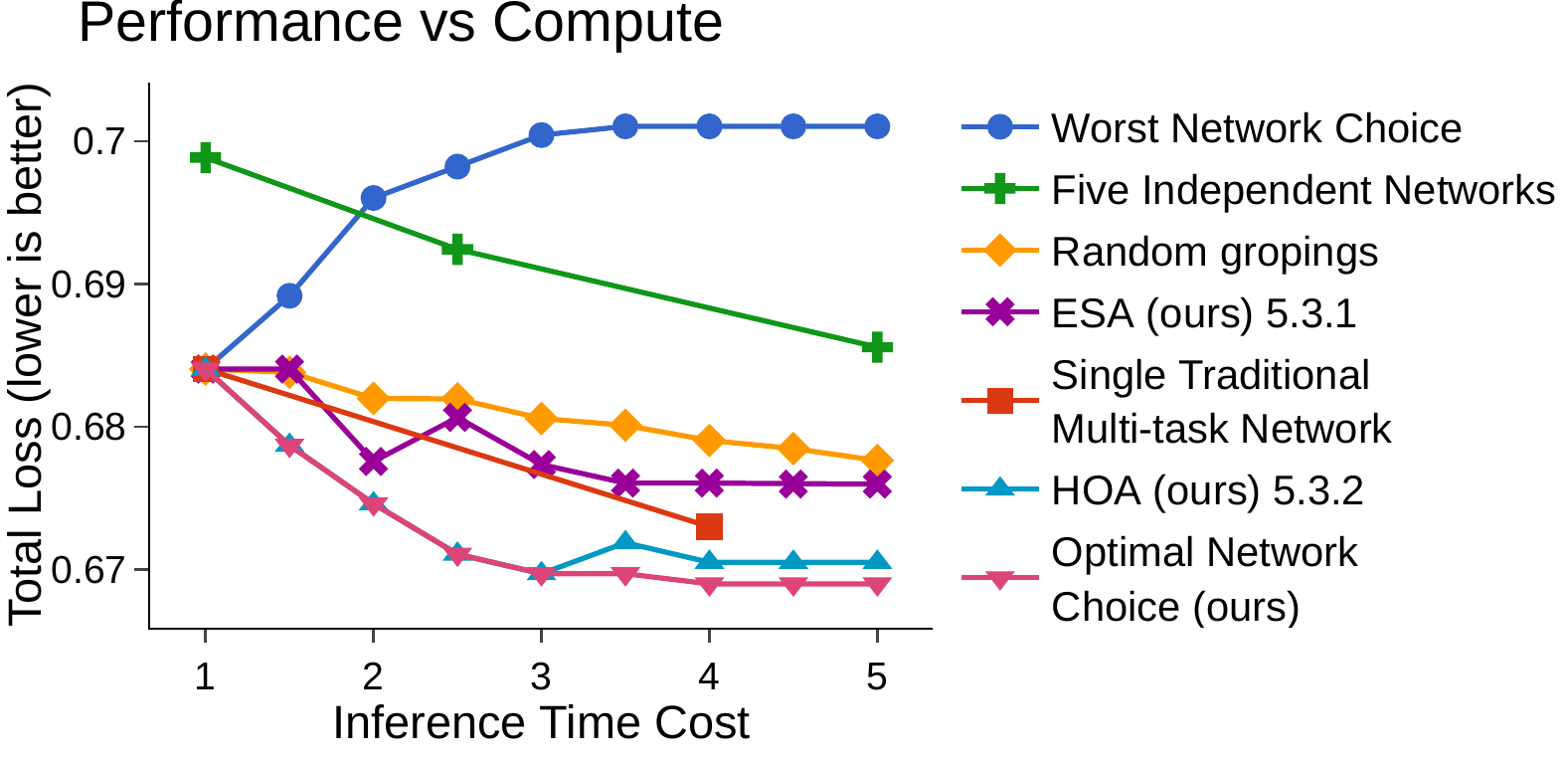}
\vspace{-10pt}
\caption{\footnotesize{\textbf{Performance/inference time trade-off in Setting 4.}}}
\vspace{-10pt}
\label{fig:alt_task_results}
\end{center}
\end{figure}

\begin{figure*}[h]
\begin{center}
\includegraphics[width=0.99\linewidth]{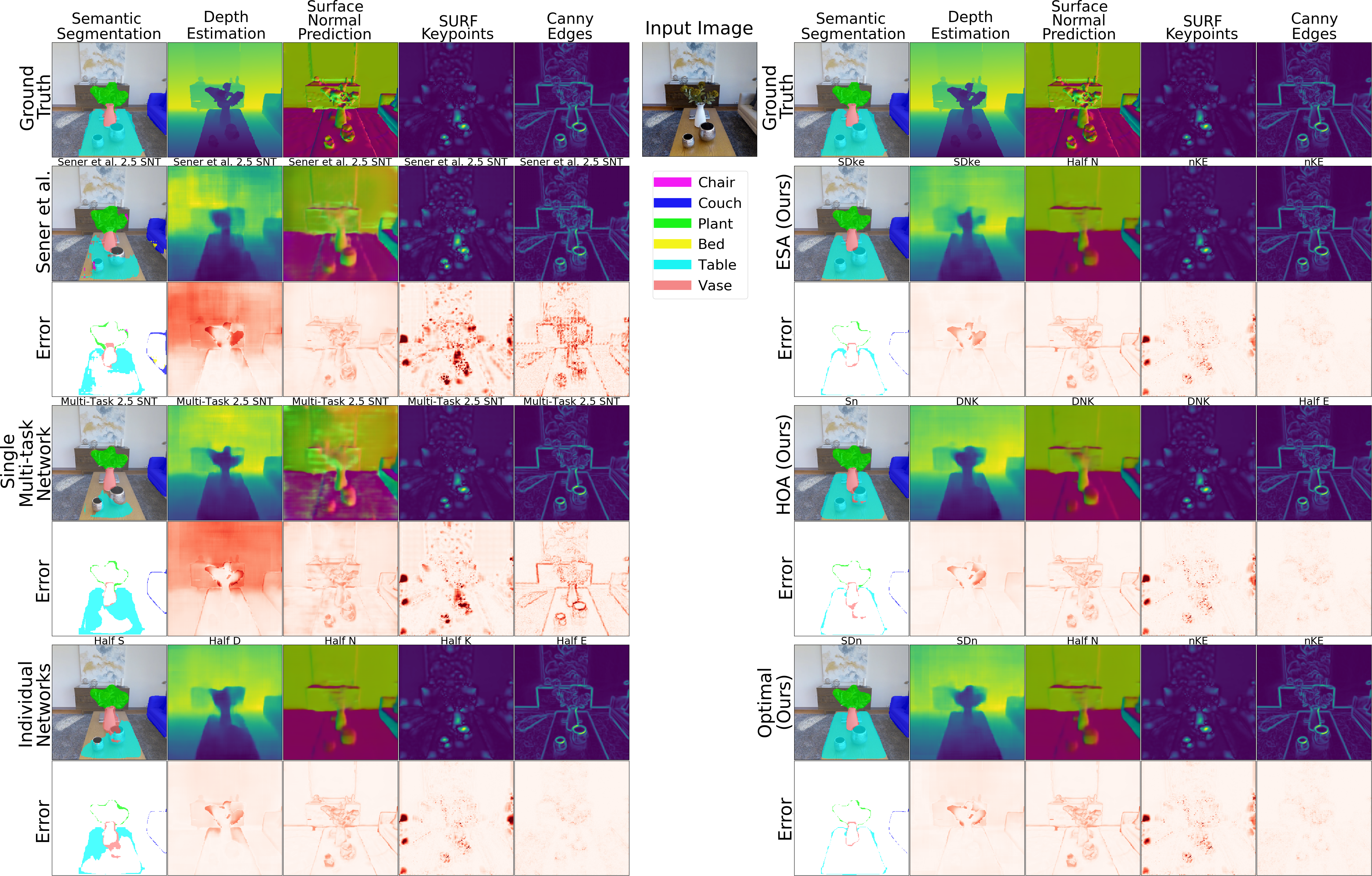}
\end{center}
\vspace{-10pt}
\caption{\footnotesize{ \textbf{Setting 1:} Qualitative results for our baselines (left) and our techniques (right). All solutions are allowed 2.5 SNT. }}
\vspace{-10pt}
\label{fig:qualitative}
\end{figure*}
\textbf{Setting 4:} In this setting, the performances of the various networks we trained were quite similar. This shows up in Figure~\ref{fig:alt_task_results} as a very small difference between the 1-SNT all-in-one, and the 5-SNT optimal solution. Since the tasks in this setting tend to cooperate well, it's not surprising that the independent training baseline is not very competitive. In fact, we see that the all-in-one network trained with 4 SNT actually outperforms the Early Stopping Approximation. Yet even in this highly cooperative scenario, our optimal solution outperforms every baseline, as does the HOA.

\textbf{Qualitative Results:} Figure~\ref{fig:qualitative} allows qualitative comparison between our methods and the baselines. We can see clear visual issues with each of the baselines that are not present in our methods. Both of our approximate methods produce predictions similar to the optimal task grouping.

\textbf{Discussion:}
In all four settings, our optimal grouping outperformed every baseline. Furthermore, the Higher Order Approximation always performed very similarly to the optimal solution, and often equally well. However, because we had five tasks, more than half of the candidate set was fully trained for this approximation. The Early Stopping Approximation was usually competitive, but the solutions tended to be worse than optimal. In principle, one could stop after training on more of the data, likely resulting in better results. We could not test ESA with other data amounts because our adaptive learning rates resulted in networks that were no longer comparable after being trained on more data.

\section{Conclusion} \label{sec:conclusion}

We describe the problem of task compatibility as it pertains to multi-task learning. We provide a computational framework for determining which tasks should be trained jointly and which tasks should be trained separately in a given setting. Our solution can take advantage of situations in which joint training is beneficial to some tasks but not others in the same group. For many use cases, this framework is sufficient, but it can be costly at training time. Hence, we offer two strategies for coping with this issue and evaluate their performance. Our methods outperform single-task networks, a multi-task network with all tasks trained jointly, as well as other baselines. We also use this opportunity to analyze how particular tasks interact in a multi-task setting and compare that with previous results on transfer learning task interactions. We find that unlike transfer task affinities, multi-task affinities depend highly on a number of factors such as dataset size and network capacity. This is another reason why a computational framework like ours is necessary for discovering which tasks should be learned together in multi-task learning.

\noindent\textbf{Acknowledgements}
We gratefully acknowledge the support of TRI~(S-2018-29), NSF grant DMS-1546206, a Vannevar Bush Faculty Fellowship, and ONR MURI~(W911NF-150100479). Toyota Research Institute (``TRI'')  provided funds to assist the authors with their research but this article solely reflects the opinions and conclusions of its authors and not TRI or any other Toyota entity.


\bibliography{mainbib}
\bibliographystyle{icml2020}

\end{document}